\documentclass{article}

\usepackage{microtype}
\usepackage{graphicx}
\usepackage{subfigure}
\usepackage{booktabs}
\usepackage{multirow}
\usepackage{makecell}
\usepackage[separate-uncertainty=true]{siunitx}
\usepackage[usenames,dvipsnames,table]{xcolor}
\usepackage{adjustbox}
\usepackage{hyperref}
\newcommand{\alignednum}[2]{%
  \hspace{-1em}\makebox[3.1em][r]{#1}{\raisebox{0ex}{\hspace{-0.15mm}$\pm$\hspace{0.2mm}#2}}
}
\newcommand{\avglen}[2]{%
  \hspace{-0.5em}\makebox[1.8em][r]{#1}{\raisebox{0ex}{\hspace{-0.15mm}$\pm$\hspace{0.2mm}#2}}%
}

\newcommand\NavyBlue{\color{NavyBlue}}

\usepackage[accepted]{2025arXiv}

\usepackage{amsmath}
\usepackage{amssymb}
\usepackage{mathtools}
\usepackage{amsthm}

\usepackage[capitalize,noabbrev]{cleveref}

\theoremstyle{plain}

\theoremstyle{definition}

\theoremstyle{remark}

\usepackage[textsize=tiny]{todonotes}
\usepackage{color}
\usepackage{soul}
\usepackage{pifont}
\newcommand{\cmark}{\ding{51}}
\newcommand{\xmark}{\textcolor{gray!40}{\ding{55}}}
\usepackage{textcomp}
\usepackage{mathtools}
\usepackage{xspace}
\usepackage{bm}

\makeatletter
\DeclareRobustCommand\onedot{\futurelet\@let@token\@onedot}
\def\@onedot{\ifx\@let@token.\else.\null\fi\xspace}
\def\eg{{e.g}\onedot}

\makeatother

\icmltitlerunning{GraspCorrect: Robotic Grasp Correction via VLM-Guided Feedback}

\begin{document}

\twocolumn[
\icmltitle{GraspCorrect: Robotic Grasp Correction \\via Vision-Language Model-Guided Feedback}

\icmlsetsymbol{equal}{*}

\begin{icmlauthorlist}
\icmlauthor{Sungjae~Lee}{equal,pos1}
\icmlauthor{Yeonjoo~Hong}{equal,pos1}
\icmlauthor{Kwang~In~Kim}{pos1,pos2}
\end{icmlauthorlist}

\icmlaffiliation{pos1}{Graduate School of AI, POSTECH}
\icmlaffiliation{pos2}{Dept. of EE, POSTECH}

\icmlcorrespondingauthor{Sungjae~Lee}{leeeesj@postech.ac.kr}
\icmlcorrespondingauthor{Yeonjoo~Hong}{yeonjooh@postech.ac.kr}
\icmlcorrespondingauthor{Kwang~In~Kim}{kimkin@postech.ac.kr}

\icmlkeywords{Machine Learning, ICML}
\vskip 0.3in
]

\printAffiliationsAndNotice{\icmlEqualContribution} 

\begin{abstract}
Despite significant advancements in robotic manipulation, achieving consistent and stable grasping remains a fundamental challenge, often limiting the successful execution of complex tasks. 
Our analysis reveals that even state-of-the-art policy models frequently exhibit unstable grasping behaviors, leading to failure cases that create bottlenecks in real-world robotic applications.
To address these challenges, we introduce GraspCorrect, a plug-and-play module designed to enhance grasp performance through vision-language model-guided feedback. GraspCorrect employs an iterative visual question-answering framework with two key components: grasp-guided prompting, which incorporates task-specific constraints, and object-aware sampling, which ensures the selection of physically feasible grasp candidates. By iteratively generating intermediate visual goals and translating them into joint-level actions, GraspCorrect significantly improves grasp stability and consistently enhances task success rates across existing policy models in the RLBench and CALVIN datasets.
\end{abstract}

\section{Introduction}
Robotic manipulation is a complex, multi-faceted task that requires the seamless integration of environmental perception, action planning, and joint actuation. From grasping delicate wine glasses to performing intricate microsurgery and assembling complex electronics, robots must exhibit precise and adaptive control to interact effectively with diverse objects and environments~\cite{Ted24}.

Recent advances in deep learning have driven significant progress in robotic manipulation, enabling the development of diverse and sophisticated robotic policies. For example, R3M uses human video data to pre-train visual representations, providing a cost-effective alternative to collecting extensive robot interaction data~\cite{NRK22}. However, its lack of explicit 3D geometric reasoning limits its precision in spatial manipulation tasks. To address this, PerAct incorporates 3D voxel patches within a transformer architecture to enhance spatial reasoning~\cite{SMF22}. Given the computational demands of high-resolution 3D features, more efficient representation techniques have also emerged~\cite{GXG23, GXGF23}. More recently, \citet{KGF24} proposed integrating diffusion policies with 3D scene representations. The resulting 3D Diffuser Actor enables the direct learning of action distribution spaces conditioned on robot states, achieving state-of-the-art performance in manipulation tasks.

Despite these advancements, the question remains: Has the manipulation problem been fully solved? Our findings suggest otherwise. Even state-of-the-art models like the 3D Diffuser Actor exhibit suboptimal performance in critical tasks. For example, in RLBench, \emph{task success} rates for scenarios such as \emph{block stacking} (68.3\%), \emph{peg insertion} (65.6\%), and \emph{shape sorting} (44.0\%) remain significantly below the benchmark average of 80\%. 

\begin{figure*}[hbt!]
    \centering
    {
    \hspace*{-0.05\textwidth}
    \begin{minipage}[t]{0.9\columnwidth}
    \begin{center}
    \includegraphics[width=0.8\linewidth]{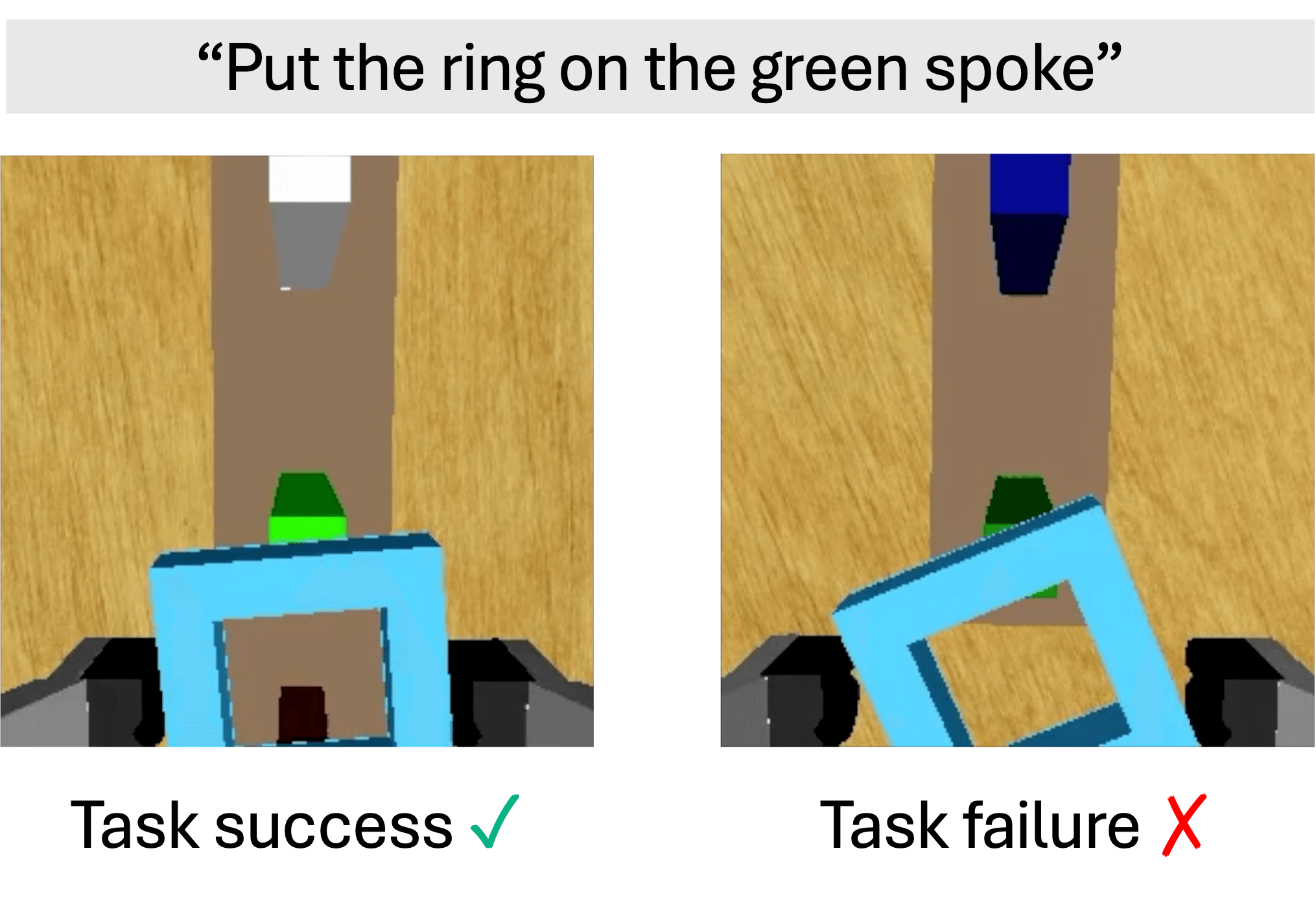}
    \end{center}
\end{minipage}
    \begin{minipage}[t]{1.1\columnwidth}
    \vspace{-3.5cm}
    \begin{center}
    \begin{small}
        \begin{tabular}{l S[table-format=2.1] S[table-format=2.1] S[table-format=2.1]}
            \toprule
            Task & 
            {Original (\%)} & 
            {Grip guided (\%)} & 
            {Improvement (\%)} \\
            \midrule
            \textit{stack blocks} & 68.3 & 88.0 & \NavyBlue 19.7 \\
            \textit{sort shape} & 44.0 & 52.0 & \NavyBlue 8.0 \\
            \textit{insert peg} & 65.6 & 92.0 & \NavyBlue 26.4 \\
            \textit{stack cups} & 47.2 & 68.0 & \NavyBlue 20.8 \\
            \textit{place cups} & 24.0 & 28.0 & \NavyBlue 4.0 \\
            \bottomrule
        \end{tabular}
    \end{small}
    \end{center}    
    \end{minipage}
    }
    \caption{Importance of precise gripper action. Left: Visualization of successful and failed cases in the RLBench \emph{insert peg} task. Right: Performance improvement on challenging RLBench tasks. By replicating demonstrated stable grasp poses up to the grasping point, we observe substantial improvements in task success rates (\%). This preliminary result highlights the significant impact of robust grasping on the overall performance of end-to-end robotic policy models.}
    \label{f:failure}
\end{figure*}

At the core of these diverse tasks lies the ability to execute stable, reliable grasps, a fundamental prerequisite for successful manipulation. We hypothesize that unstable grasping is a persistent bottleneck across these tasks. To test this, we conducted a preliminary experiment where stable grasp trajectories were provided up to the grasp point, allowing the policy model to predict actions thereafter. This intervention yielded significant improvements, with task success rates increasing by up to 26.4\%~(\cref{f:failure}).

To address these challenges, two main strategies have been proposed. One direction focuses on creating extensive datasets of robot demonstrations that encompass a wide range of object properties, grasp types, and environmental conditions.
However, such approaches are resource-intensive, requiring substantial computational power for retraining, and are often limited by scope of the datasets~\cite{PSD24}. Alternatively, pre-trained grasping foundation models like GraspNet~\cite{FWG20} and Contact-GraspNet~\cite{SMT21} generate efficient grasp pose distributions directly from 3D point clouds. While effective in dense point cloud, these models face performance degradation when applied to cost-efficient, sparse 3D representations. Furthermore, their reliance on scenes observed during training makes it difficult to generalize to unseen objects and task categories.

A promising alternative lies in the zero-shot capabilities of Vision-Language Models (VLMs), which excel in scene understanding, reasoning, and generating descriptive text~\cite{YLL24}. Recent studies demonstrated their effectiveness in robotic manipulation and long-horizon planning tasks~\cite{HLH24, KDJ24}. Trained on diverse, large-scale datasets, VLMs offer the adaptability required for robust performance across varying domains. 

We hypothesize that integrating VLMs into the grasp correction process can enhance performance by analyzing visual context and suggest adjustments, without requiring extensive retraining or large-scale data collection.

Building on these insights, we propose GraspCorrect, a plug-and-play module specifically designed to enhance grasp reliability in robotic manipulation policies. GraspCorrect operates through a three-stage pipeline. First, it identifies stable grasp positions using VLM guidance that integrates semantic and geometric considerations. Next, it synthesizes a goal-state image via image composition, translating VLM insights into actionable visual objectives. Finally, Goal-Conditioned Behavioral Cloning (GCBC) converts these visual goals into precise joint-level actions, ensuring accurate execution of the intended grasp. 

This architecture-agnostic design allows for seamless integration with existing manipulation policies, enabling grasp refinement without compromising core functionality or requiring extensive retraining. By effectively predicting corrective actions during the critical grasping phase, GraspCorrect significantly improves grasp reliability, achieving state-of-the-art performance in robotic manipulation.

\section{Related Work}
\paragraph{Policy Learning for Robotic Manipulation:}
Policy learning enables robots to autonomously develop complex skills for interacting with diverse environments. While Reinforcement Learning (RL) has been widely used to learn optimal policies through interaction with the environment, it often suffers from data inefficiency and requires extensive exploration, limiting its real-world applicability~\cite{TAH24}. 
Imitation Learning (IL) leverages expert demonstrations to train policies, reducing the exploration burden. Nonetheless, IL faces challenges with generalization to unseen scenarios~\cite{MXM20}.

Recent advancements in policy learning architectures have gained increasing attention. Transformer-based architectures have demonstrated the ability to capture long-range dependencies and contextual relationships in manipulation tasks~\cite{SMF22,GXG23,GBX24}. Similarly, diffusion models have been explored to represent complex action distributions~\cite{KGF24, BNA24}. Despite these developments, robust manipulation remains a significant challenge, particularly in tasks requiring precise grasping~\cite{GBX24}. This highlights the importance of enhancing grasping capabilities within policy learning to improve overall manipulation.

A closely related approach is SuSIE~\cite{BNA24}, which employs an image-editing diffusion model to generate intermediate sub-goal frames for long-horizon manipulation tasks. However, GraspCorrect is motivated by a key insight: correct grasping serves as a crucial milestone for the success of manipulation tasks. As such, GraspCorrect explicitly uses VLMs to assess the likelihood of a correct grasp and to generate corresponding images, a capability that SuSIE lacks. Our empirical observations show that SuSIE struggles to generate appropriate grasp poses, despite being fine-tuned on a comprehensive dataset that combines both simulation and real-world scenarios (see~\cref{f:edit}).

\paragraph{Vision Language Models in Robotics:}
Vision Language Models (VLMs) have significantly advanced robotic interaction through their capabilities in advanced scene understanding, task planning, and descriptive captioning. For example, PaLM-E~\cite{DXS23} integrates language models with visual and motor capabilities, while RT-1~\cite{BBC23} introduces efficient tokenization techniques for processing high-dimensional camera inputs and motor commands, enabling real-time control.

Despite these advancements, VLMs still struggle to bridge the gap between high-level understanding and low-level control, particularly in maintaining temporal consistency and spatial precision. Even with recent architectures like RT-2~\cite{BBCC23}, CoPa~\cite{HLH24}, and EmbodiedGPT~\cite{MZH24}, achieving precise manipulation remains difficult. Our work complements these advances by focusing on grasp correction, using VLMs' scene understanding and commonsense reasoning capabilities.

PIVOT~\cite{NXY24} introduces a visual prompting approach for VLMs, reformulating complex manipulation tasks as iterative visual question-answering (VQA). In each cycle, the image is annotated with visual representations of proposals, allowing the VLM to iteratively refine and select the most suitable option. While we adopt this iterative selection scheme, we observe that it lacks an inherent understanding of physical feasibility and constraints within the visual scene. 

As shown in \cref{s:ablation}, GraspCorrect effectively addresses these limitations by better constraining sampling strategies and incorporating physical interaction principles.

\begin{figure*}
\begin{center}
\includegraphics[width=\textwidth]{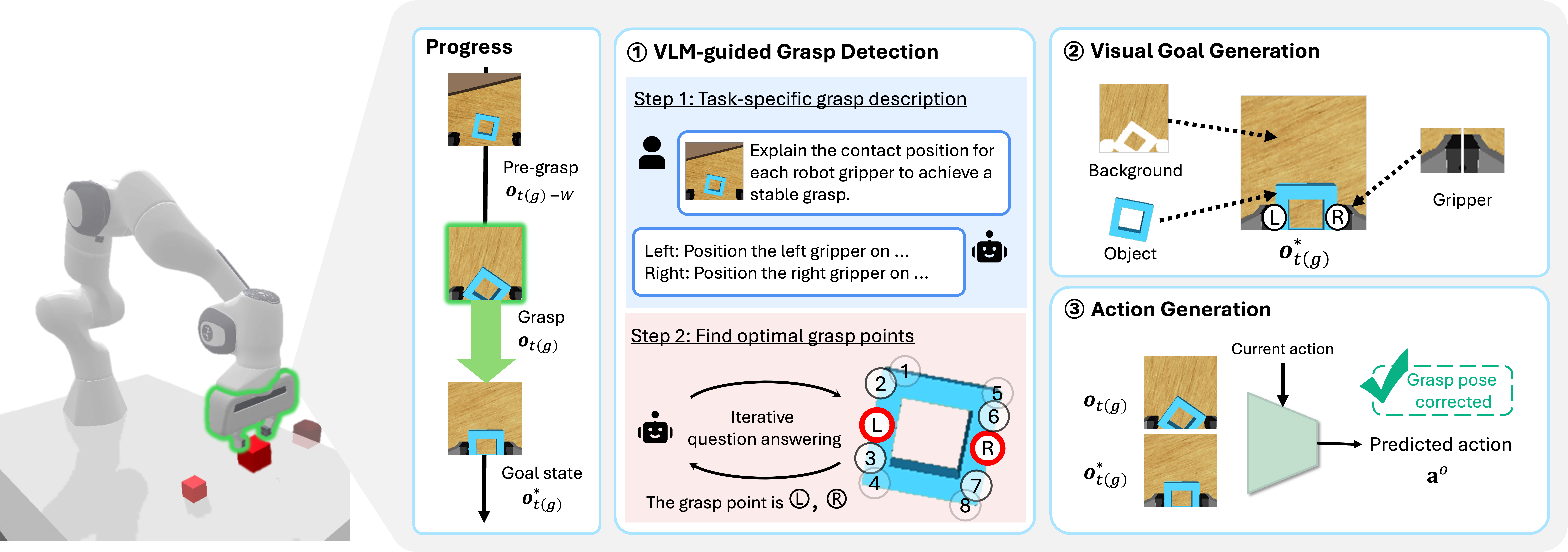}
\caption{Overview of the GraspCorrect process. This module enhances robotic manipulation by establishing a stable grasp as a critical milestone. In the Grasp Detection stage, task-specific VLM guidance predicts the desired gripper positioning through an iterative question-answering process. The Visual Goal Generation stage then synthesizes a goal-state image via image composition, representing the ideal grasp configuration. Finally, the Action Generation stage predicts and executes corrective actions, improving grasping reliability.}
\label{f:main}
\end{center}
\end{figure*}

\section{Robotic Grasp Correction Framework}
We consider robotic manipulation learning, where a policy model learns from demonstration trajectories $\{(\mathbf{o}_1, \mathbf{a}_1), (\mathbf{o}_2, \mathbf{a}_2),...\}$ paired with a textual task instruction $l$. Each observation $\mathbf{o}_t$ at timestep $t$ consists of RGB-D images, and each action 
\begin{align}
\label{e:adecompos}
\mathbf{a}_t=[(\mathbf{a}^{p}_t)^\top,(\mathbf{a}^{r}_t)^\top,(\mathbf{a}^{s}_t)^\top]^\top\in \mathbb{R}^{8}
\end{align}
specifies the end-effector pose through three components: position $\mathbf{a}^{p}_t \in \mathbb{R}^3$, rotation (quaternion) $\mathbf{a}^{r}_t \in \mathbb{R}^4$, and binary gripper state $\mathbf{a}^{s}_t \in \{0,1\}$. 

The policy model aims to learn a mapping that predicts an appropriate action $\mathbf{a}$ given the current observation $\mathbf{o}$. GraspCorrect acts as a plug-and-play module for existing policy models, activating at the grasping moment $t(g)$ when the gripper contacts the target object specified in $l$. Using the current grasp pair $(\mathbf{o}_{t(g)},\mathbf{a}_{t(g)})$ and a pre-grasp observation $\mathbf{o}_{t(g) - W}$ within a time window $W$ (see~\cref{s:graspdetection}), the module predicts a corrective end-effector grasp pose $\mathbf{a}^o$ for improved execution. 

GraspCorrect operates through three stages. First, in the \emph{(VLM-guided) Grasp Detection} stage, it identifies stable grasp positions using insights from VLM. In the \emph{(Visual) Goal Generation} stage, it generates a visual objective in image form. Finally, in the \emph{Action Generation} stage, it translates the visual goal into precise joint-level actions. \cref{f:main} illustrates the overall grasp correction process. 

\begin{figure}[hbt!]
\vspace{3mm}
\begin{center}
\includegraphics[width=\columnwidth]{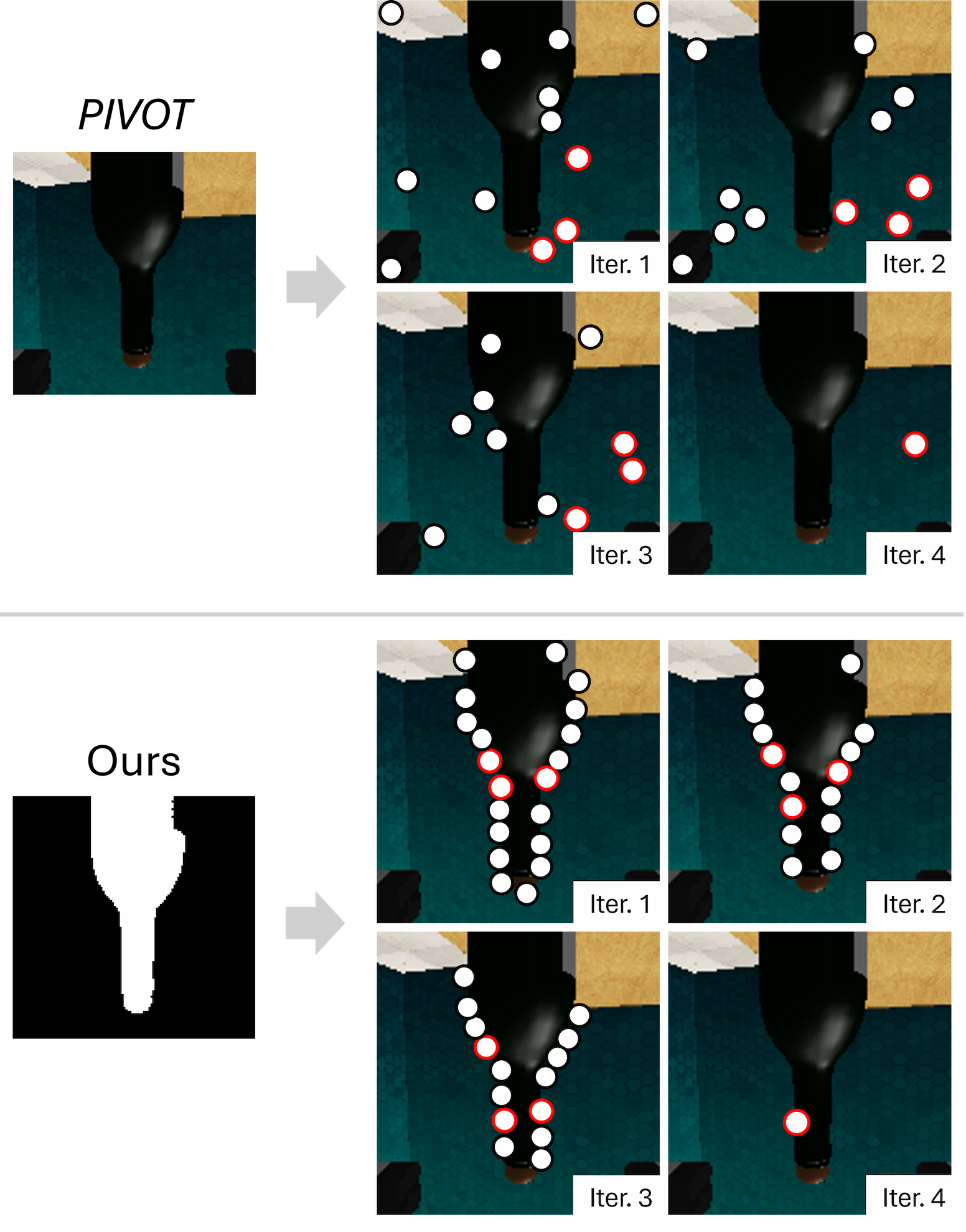}
\caption{Visualization of iterative grasp point refinement using PIVOT~\cite{NXY24} (top) and our method (bottom). The circles represent grasp candidates sampled by each algorithm, with red circles indicating those selected for the next sampling stage. Due to its lack of target-specific contextualization, PIVOT often predicts grasp points that fail to make contact with the object. In contrast, our method ensures all selected grasp locations are physically viable. The left and right gripper positions are aligned within the camera's image pane, making it sufficient to generate grasp points near the image boundaries (see~\Cref{f:failure}, left).}
\label{f:vlm}
\end{center}
\end{figure}

\subsection{VLM-guided Grasp Detection}
\label{s:graspdetection}
This stage translates the current observation $\mathbf{o}_{t(g)}$ and task description $l$ into task-oriented contact points $\mathbf{a}^p_{t(g)}$ for robotic grippers to ensure stable grasping. Leveraging VLMs, this task can be framed as spatial Visual Question Answering (VQA), which extends traditional VQA tasks (\eg, identifying objects or attributes; ``What color is the car?'') to include spatial reasoning, such as determining where a robot should grasp an object for a stable lift. 

Pre-trained VLMs provide a rich repository of commonsense knowledge for this task. However, their direct application to spatial reasoning presents two main challenges. First, VLMs are optimized for generating textual outputs, making them unsuitable for producing continuous values like coordinates or actions. Second, even state-of-the-art VLMs struggle with complex spatial reasoning~\cite{WMS24,CXK24,TQW24}.

To overcome these limitations, we adopt an iterative VQA approach that progressively refines grasp candidates rather than attempting to generate precise spatial coordinates directly. Building on the iterative refinement strategy of PIVOT~\cite{NXY24}, we introduce two key improvements: (1) grasp-guided prompting, which incorporates task-specific constraints, and (2) object-aware sampling, which ensures that generated grasp candidates are physically feasible.

Our approach starts with a top-down 2D observation $\mathbf{o}_{t(g) - W}^{\text{Top}}$, captured $W$ frames before the grasping moment. This earlier frame provides a comprehensive view of object’s geometry, as the close-up grasp pose at $t(g)$ may only partially capture the object. The time window size $W$ is fixed at 10. Using curated prompts tailored to task requirements (see~\cref{sss:prompt}), the VLM generates a textual description of stable grasp configurations, which serves as a prior for the iterative refinement process. 

To ensure precise targeting, we use LangSAM,\footnote{https://github.com/luca-medeiros/lang-segment-anything} a zero-shot text-to-segmentation-mask framework that combines GroundingDINO~\cite{LZR25} and Segment-Anything~\cite{KMR23}. This segmentation step restricts grasp proposals to the actual object, avoiding hallucinations that could target background elements.

Grasp candidates are initially sampled along the object's contour~(\cref{f:vlm}, circles). The VLM evaluates these points to identify \emph{promising candidates} (red circles) likely to support stable grasping. New candidates are then generated by sampling from a 1D Gaussian distribution centered around these promising points along the object’s contour. The number of iterations $T$ is fixed at 4, and in the final iteration, a single candidate is selected. A detailed description of this process can be found in~\cref{s:vlmguidedgraspdetails}.

\subsection{Visual Goal Generation}
This stage synthesizes a target grasp pose image $\mathbf{o}_{t(g)}^*$ that depicts the robotic grippers (left and right), the target object, and their spatial relationships, based on the input observations $\{\mathbf{o}_{t(g)},\mathbf{o}_{t(g) - W}\}$, and the grasp points identified by the Grasp Detection stage.

The process starts by restoring the occluded background regions using the LaMa inpainting model~\cite{SLM22} to create a complete background image. The composite image is then constructed by blending the restored background, gripper, and transformed foreground object. Object alignment with the gripper is achieved via conventional image transformations (rotations and translations) guided by the contact point information from the Grasp Detection stage. The resulting goal-state image provides a realistic representation of the desired grasp pose and serves as the foundation for the subsequent Action Generation step.

\subsection{Action Generation}
To achieve low-level joint actuation, we adopt a Goal-Conditioned Behavior Cloning (GCBC) framework. As a form of imitation learning, behavior cloning trains an agent to replicate expert demonstrations by minimizing the discrepancy between predicted and observed expert actions.
Following~\cite{WBL23}, we implement this using Denoising Diffusion Probabilistic Models (DDPM)~\cite{HJA20}, which iteratively refines a Gaussian noise distribution into a data-generating distribution.

Our GCBC policy model $\pi_\phi$ comprises a ResNet-34 encoder followed by 3-layer Multi-Layer Perceptron (MLP), parameterized by weights $\phi$. Since observation images are captured from an egocentric top-down perspective, we enhance spatial awareness by incorporating the current action state as a conditioning variable. This facilitates smooth integration of generated output actions into the ongoing trajectory. 

The DDPM training loss is formulated as:
\begin{align}
\mathcal{L}(\phi) =  &\mathbb{E}_{\bm{\epsilon},s,(\mathbf{a}_{t}, \mathbf{a}_{t}^{*}, \mathbf{o}_{t}, \mathbf{o}_{t}^{*})\sim \mathcal{D}} \big[ \lambda\|\bm{\epsilon}^{p}(s) - \bm{\epsilon}^{p}_\phi(\mathbf{a}_{t}, \mathbf{o}_{t}, \mathbf{o}_{t}^{*}, s) \|^2\nonumber \\
&+\|\bm{\epsilon}^{r}(s) - \bm{\epsilon}^{r}_\phi(\mathbf{a}_{t}, \mathbf{o}_{t}, \mathbf{o}_{t}^{*}, s) \|^2\big],
\label{e:loss}
\end{align}
where $s$ is the diffusion time step, and $t$ represents the grasp time $t(g)$. Here, $\mathbf{o}_{t}$ is the observation image, $\mathbf{a}_t$ is the action vector (\cref{e:adecompos}), $\mathbf{o}_t^{*}$ is the goal image, and $\mathbf{a}_t^{*}$ is the expert action. The noise vectors $\bm{\epsilon}^{p}_\phi$ and $\bm{\epsilon}^{r}_\phi$, corresponding to position and rotation, respectively, are predicted by $\pi_\phi$ to approximate the true noise terms $\bm{\epsilon}^{p}(s)$ and $\bm{\epsilon}^{r}(s)$ associated with $\mathbf{a}_{t}^{*}$. 
The weighting hyperparameter $\lambda$ is set to 0.2 based on validation (see~\cref{s:ablation} for details). The expectation in $\mathcal{L}$ is taken over $\mathbf{a}$ and $\bm{\epsilon}$, where $\mathbf{a}$ comprises $\mathbf{a}^p$ and $\mathbf{a}^r$, and $\bm{\epsilon}$ includes $\bm{\epsilon}^{p}$ and $\bm{\epsilon}^{r}$.

Training data $\mathcal{D}$ is generated within each benchmark environment by systematically perturbing ground-truth grasp poses. Further details on the policy model and data generation process can be found in~\cref{ss:action}.

\begin{figure}[t!]
\vspace{3mm}
\begin{center}
\hspace{-0.5em}
\includegraphics[width=1.01\columnwidth]{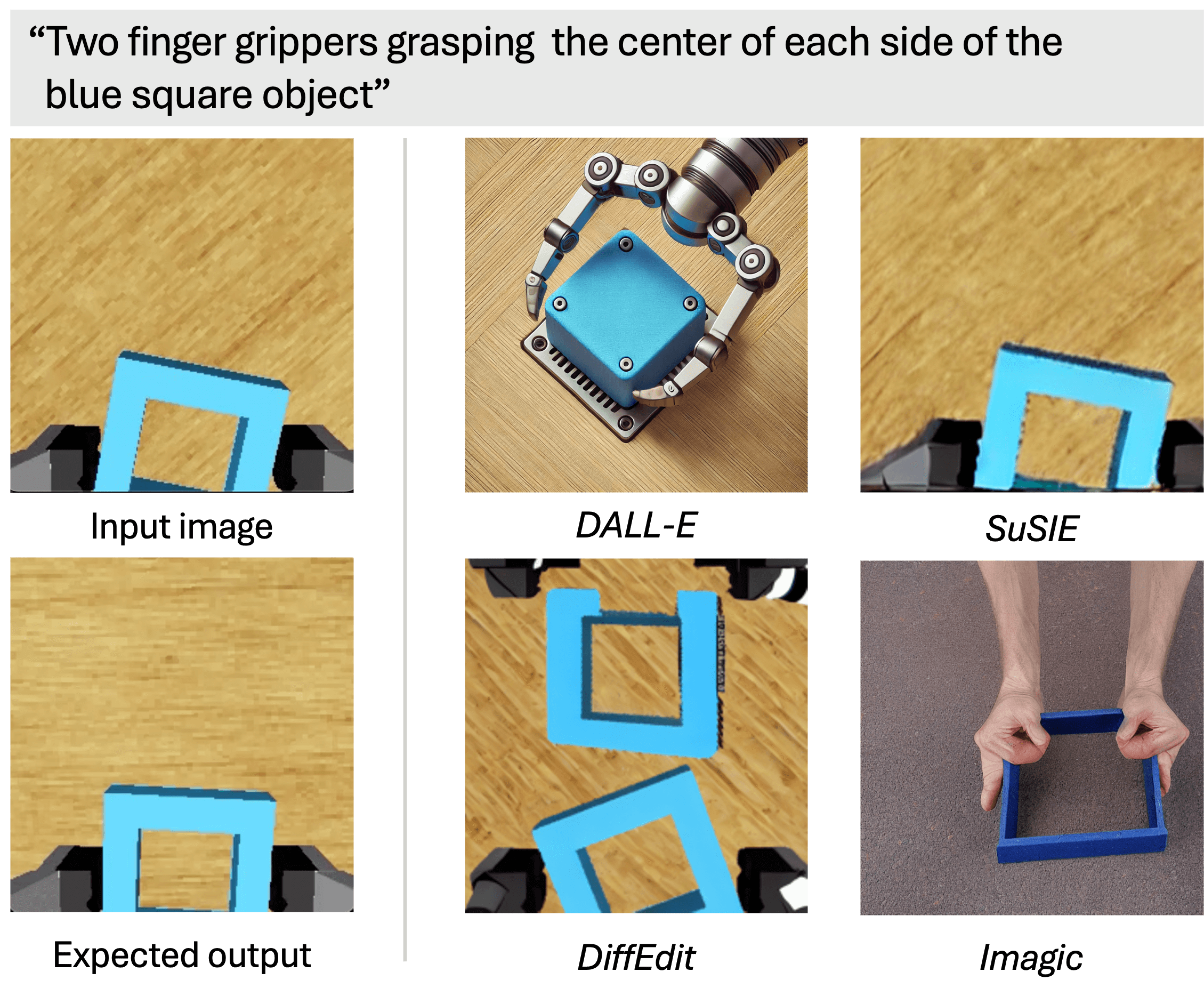}
\caption{Evaluation of diffusion-based models for generating goal-state images in robotic manipulation tasks. The input image (top-left) shows the initial gripper configuration approaching a blue square object, while the expected output (bottom-left) represents the ground-truth stable grasp pose from a successful RLBench \emph{insert peg} demonstration. Existing models struggle to accurately capture the required details, spatial arrangements, and contextual relevance essential for precise robotic grasping.}
\label{f:edit}
\end{center}
\end{figure}

\subsection{Discussion}
\paragraph{Complementary Roles of VLMs and Behavioral Control:}
Our approach combines a VLM for grasp detection with GCBC for action generation, recognizing the limitations of VLMs in directly synthesizing precise grasping actions. While VLMs excel in scene understanding and high-level planning, they struggle with the fine-grained control required for embodied manipulation. 

In our preliminary experiments, directly using VLMs for action prediction ($\mathbf{a} \in \mathbb{R}^{8}$) based on current observations, actions, and task descriptions often resulted in unrealistic and physically implausible outputs. This validates our decision to partition the manipulation pipeline, using VLMs for perception and planning while relying on a specialized GCBC module for precise control.

\paragraph{Advantages of Image-Based Intermediate Representations:}
GraspCorrect uses images as the intermediate goal representation. This decision is grounded in several key advantages. First, visual representations capture rich spatial and contextual information that might be lost or ambiguous in text-based descriptions. Images naturally encode crucial manipulation-relevant features such as spatial relationships, object orientations, and grasp configurations in a concrete, unambiguous manner. 

Second, VLMs have been extensively trained on large-scale visual data, making them particularly adept at processing and reasoning about image-based information. This alignment enables our system to fully exploit VLMs' advanced visual understanding and reasoning capabilities while maintaining a clear and interpretable interface for high-level decision-making. 

Third, recent successes in robotic manipulation using synthesized goal images, such as SuSIE~\cite{BNA24} and GR-MG~\cite{LWH25}, further demonstrate the effectiveness of this approach.

However, image-based representations also present certain limitations, particularly in handling occlusions and capturing dynamic physical properties. Future work could explore incorporating additional modalities, such as 3D point clouds or force feedback, to provide a richer and more comprehensive representation of the goal state.

\paragraph{Alternative Visual Goal Generation Strategies:}
The expected goal-state grasp pose typically requires minor adjustments from the current grasp pose provided by a pre-trained policy model. While path-planning algorithms like Rapidly-exploring Random Trees (RRT)~\cite{Lav98} might seem applicable, they are unsuitable for this context as they require exact destination coordinates, which are challenging to derive due to mismatches between the egocentric robotic camera and the robot's coordinate system. Consequently, pinpointing the precise coordinate indicated by VLMs' responses becomes challenging.

An alternative approach involves using image-editing or image-generation diffusion models prompted by the VLM-generated grasp descriptions. We tested four such models in RLBench: DALL-E (image generation)~\cite{RPG21} , SuSIE~\cite{BNA24}, DiffEdit~\cite{CVS23}, and Imagic (image editing)~\cite{KZL23}. Following SuSIE's approach, we employed a fine-tuned version of InstructPix2Pix~\cite{BHE23} specifically adapted for manipulation tasks. 

As shown in~\cref{f:edit}, these models often fail to generate accurate and reliable results. For instance, DALL-E tends to generate overly intricate robotic grippers that deviate significantly from the actual gripper design. DiffEdit and SuSIE misrepresent the square’s orientation, failing to align it with the expected grasp pose, while Imagic introduces unrealistic human fingers.

In contrast, a simple image blending technique proves highly effective for generating accurate and realistic composite images of the goal state. This approach preserves the structural integrity of the target object and maintains the spatial relationships critical for successful manipulation.

\setlength{\tabcolsep}{2pt} 
\begin{table*}[hbt!]
\caption{Performance across different manipulation frameworks on 18 tasks from the RLBench dataset. For each example in each task, the initial position (and orientation) of the target object are randomly generated three times. The mean task success rate (\%) $\pm$ standard deviations is computed across all examples and random object placement combinations within each task. The dash symbol (-) indicates that GraspCorrect remains inactive for tasks that do not involve object grasping.}
\label{t:result}
\vspace{1mm}
\begin{center}
\begin{small}
\resizebox{16.8cm}{!}{
\begin{tabular}{l c c c c c|c c c c}
\toprule
 & {\makecell{\textit{stack}\\\textit{blocks}}} & {\makecell{\textit{sort}\\\textit{shape}}} & {\makecell{\textit{insert}\\\textit{peg}}} & {\makecell{\textit{stack}\\\textit{cups}}} & {\makecell{\textit{place}\\\textit{cups}}} &
 {\makecell{\textit{sweep to}\\\textit{dustpan}}} &
 {\makecell{\textit{turn}\\\textit{tap}}} &
 {\makecell{\textit{put in}\\\textit{drawer}}} &
   {\makecell{\textit{close}\\\textit{jar}}} \\
\midrule
PerAct  & \alignednum{29.3}{6.1} & \alignednum{17.3}{2.3}  & \alignednum{5.3}{2.3} & \alignednum{0.0}{0.0} & \alignednum{1.3}{2.3} & \alignednum{42.7}{8.3} & \alignednum{82.7}{4.6} & \alignednum{53.3}{2.3} & \alignednum{50.7}{2.3}\\
 + Ours & \alignednum{33.3}{6.1} & \alignednum{21.3}{2.3} & \alignednum{8.0}{6.9} & \alignednum{0.0}{0.0} & \alignednum{1.3}{2.3} & \alignednum{42.7}{8.3} & \alignednum{84.0}{4.0} & \alignednum{64.0}{6.9} & \alignednum{74.7}{2.3}\\
\midrule
Act3D   & \alignednum{4.0}{4.0} & \alignednum{36.0}{4.0}& \alignednum{16.0}{0.0} & \alignednum{6.7}{2.3} & \alignednum{9.3}{6.1} & \alignednum{88.0}{10.6} & \alignednum{90.7}{6.1} & \alignednum{93.3}{4.6} & \alignednum{96.0}{0.0}\\
 + Ours   & \alignednum{6.7}{2.3} & \alignednum{37.3}{6.1} & \alignednum{24.0}{4.0} & \alignednum{6.7}{2.3} & \alignednum{9.3}{6.1} & \alignednum{88.0}{10.6} & \alignednum{94.7}{4.6} & \alignednum{96.0}{2.3} & \alignednum{98.7}{2.3}\\
\midrule
3D Diff. Act. & \alignednum{66.7}{4.5} & \alignednum{46.0}{4.0} & \alignednum{65.3}{2.3} & \alignednum{40.0}{4.0} & \alignednum{21.3}{6.1} & \alignednum{89.3}{1.5} & \alignednum{97.3}{2.3} & \alignednum{90.7}{2.3} & \alignednum{98.7}{2.3} \\
 + Ours & \alignednum{76.0}{4.0} & \alignednum{60.0}{4.0} & \alignednum{82.7}{2.0} & \alignednum{65.3}{6.1} & \alignednum{45.3}{16.2} & \alignednum{89.3}{1.5} & \alignednum{98.7}{2.3} & \alignednum{94.0}{2.8} & \alignednum{100.0}{0.0}\\
\midrule
RVT-2   & \alignednum{78.7}{4.6}  & \alignednum{34.7}{9.2} & \alignednum{48.0}{4.0} & \alignednum{73.3}{8.3}  &  \alignednum{44.0}{4.0} & \alignednum{100.0}{0.0} & \alignednum{98.7}{2.3} & \alignednum{97.3}{2.3} & \alignednum{100.0}{0.0}\\
 + Ours   & \alignednum{82.7}{4.6} & \alignednum{49.3}{6.1} & \alignednum{54.7}{4.6} & \alignednum{82.7}{2.3} & \alignednum{61.3}{6.1} & \alignednum{100.0}{0.0} & \alignednum{98.7}{2.3} & \alignednum{100.0}{0.0} & \alignednum{100.0}{0.0}\\
\midrule
 & {\makecell{\textit{screw}\\\textit{bulb}}} & {\makecell{\textit{place}\\\textit{wine}}} & {\makecell{\textit{meat off}\\\textit{grill}}} &  {\makecell{\textit{put in}\\\textit{cupboard}}} & {\makecell{\textit{open}\\\textit{drawer}}} &
 {\makecell{\textit{drag}\\\textit{stick}}} &
 {\makecell{\textit{put in}\\\textit{safe}}} &
 {\makecell{\textit{push}\\\textit{buttons}}} &
  {\makecell{\textit{slide}\\\textit{block}}} \\
\midrule
PerAct   & \alignednum{25.3}{2.3} & \alignednum{48.0}{6.9} & \alignednum{70.7}{2.3} & \alignednum{33.3}{8.3} & \alignednum{50.7}{2.3} & \alignednum{89.3}{6.1} & \alignednum{80.0}{6.9} & \alignednum{94.7}{2.3} & \alignednum{80.0}{13.9}\\
 + Ours    &  \alignednum{25.3}{2.3} & \alignednum{49.3}{6.1} & \alignednum{80.0}{0.0} & \alignednum{36.0}{8.0} & \alignednum{96.0}{0.0} & \alignednum{90.7}{6.1} & \alignednum{80.0}{6.9} & - & -\\
\midrule
Act3D    & \alignednum{33.3}{8.3} & \alignednum{60.0}{8.0} & \alignednum{97.3}{2.3} & \alignednum{66.7}{4.6} & \alignednum{85.3}{6.1} & \alignednum{66.7}{2.3} & \alignednum{98.7}{2.3} & \alignednum{93.3}{2.3} & \alignednum{93.3}{4.6}\\
 + Ours    & \alignednum{45.3}{2.3} & \alignednum{60.0}{8.0} & \alignednum{97.3}{2.3} & \alignednum{69.3}{2.3} & \alignednum{93.3}{8.3} & \alignednum{68.0}{0.0} & \alignednum{98.7}{2.3} & - & -\\
\midrule
3D Diff. Act. & \alignednum{69.3}{2.3} & \alignednum{88.0}{8.0} & \alignednum{90.7}{6.1} & \alignednum{78.7}{2.3} & \alignednum{90.7}{4.6} & \alignednum{98.7}{2.3} & \alignednum{97.3}{2.3} & \alignednum{97.3}{2.3} & \alignednum{98.7}{2.3}\\
 + Ours & \alignednum{86.7}{2.3} & \alignednum{93.3}{2.3} & \alignednum{97.3}{4.6} & \alignednum{90.7}{2.3} & \alignednum{98.7}{2.3} & \alignednum{98.7}{2.3} & \alignednum{97.3}{2.3} & - & -\\
\midrule
RVT-2    & \alignednum{92.0}{0.0} & \alignednum{89.3}{4.6} & \alignednum{97.3}{2.3} & \alignednum{68.0}{4.0} & \alignednum{72.0}{6.9} & \alignednum{100.0}{0.0} & \alignednum{97.3}{4.6} & \alignednum{100.0}{0.0} & \alignednum{93.3}{4.6}\\
 + Ours     & \alignednum{92.0}{0.0} & \alignednum{94.7}{2.3} & \alignednum{98.7}{2.3} & \alignednum{72.0}{0.0} & \alignednum{88.0}{4.0} & \alignednum{100.0}{0.0} & \alignednum{97.3}{4.6} & - & -\\
\bottomrule
\end{tabular}
}
\end{small}
\end{center}
\end{table*}

\section{Experiments}
To assess the effectiveness of our GraspCorrect module, we conducted experiments on RLBench~\cite{JMR20} and CALVIN~\cite{MHE22}. For all baseline models, we used the pre-trained weights provided by the respective authors. VLM guidance was provided by ChatGPT-4o~\cite{Open25} in its standard form without modifications. All experiments followed the default setup of each benchmark, using the Franka Panda Arm with a parallel-jaw gripper (see~\cref{f:main}). While our evaluation focuses on this specific configuration, the framework is adaptable to other VLMs and vision-based robotic manipulation tasks.

\paragraph{RLBench:}
RLBench is a widely used benchmark in robotic learning research~\cite{SMF22,GXGF23,KGF24,GBX24}, providing 100 diverse manipulation tasks that simulate real-world scenarios. We focus on 18 fundamental manipulation tasks that are broadly recognized within the robotics community, following the evaluation framework proposed by~\cite{SMF22}. Among these, the first five tasks listed in~\cref{t:result}, in raster order, represent particularly challenging scenarios where state-of-the-art methods consistently achieve success rates below 80\%. For baseline comparisons, we evaluated four manipulation models: PerAct~\cite{SMF22}, Act3D~\cite{GXGF23}, 3D Diffuser Actor~\cite{KGF24} and RVT-2~\cite{GBX24}.

\Cref{t:result} summarizes the results. While 3D Diffuser Actor and RVT-2 demonstrate strong baseline performance, they occasionally struggle during critical grasping moments. Even minor inaccuracies in grasp execution can lead to task failures, despite the preceding execution sequence being correct (see task failure example in \cref{f:failure}).

Integrating GraspCorrect with these architectures led to substantial performance improvements, averaging 18.3\% and 5.5\%, respectively, across evaluation tasks. By providing targeted refinement during the grasping phase, GraspCorrect effectively mitigates these models' primary performance bottleneck while preserving their advanced trajectory generation and dynamic control capabilities. This highlights how precise intervention at key manipulation stages can further enhance reliability, even in already robust policies.

For PerAct and Act3D, performance gains were consistently observed, but they were more modest. We attribute this to two key limitations. First, these models struggle with precise height prediction for keypoints, often resulting in complete grasp failures before GraspCorrect can intervene. Since our module activates specifically during the grasping phase, it cannot compensate for cases where the initial approach fails to position the gripper within a viable correction range. 

Second, these models exhibit multi-task performance limitations that extend beyond grasping. While GraspCorrect improves initial grasp stability, failures still occur in subsequent manipulation stages, particularly in object release orientation and adherence to language-conditioned constraints.

Overall, GraspCorrect often significantly enhances task performance, with particularly strong improvements in the most challenging tasks (the first five in \cref{t:result}). This supports our hypothesis that unstable grasping remains a major bottleneck in robotic manipulation. \Cref{f:visualization} provides a comparative visualization of manipulation trajectories, illustrating the impact of GraspCorrect on task completion. 

Performance gains are more moderate in certain cases (last four columns in \cref{t:result}). In these tasks, baseline models already exhibited near-perfect performance, or the tasks did not involve object grasping, meaning our module was not invoked. Importantly, GraspCorrect did not degrade performance in any case.

The results also highlight areas for improvement. GraspCorrect's impact is limited in cases where failures occur earlier in the manipulation pipeline, such as approach planning or object localization errors. Additionally, since our module does not fully modify the action policy, it does not address post-grasp actions, including object placement orientation or adherence to task constraints. Future work could extend GraspCorrect to intervene at earlier manipulation stages or integrate it within a broader policy refinement framework to enhance full-sequence task execution.

\Cref{f:graspexample} shows representative examples across four manipulation scenarios. GraspCorrect consistently generates grasp poses closely aligned with ground-truth demonstrations. In contrast, Contact-GraspNet struggles to generalize to unseen object-task combinations beyond its training settings.

\paragraph{CALVIN:} 
The CALVIN dataset provides an environment for learning long-horizon language-conditioned tasks, encompassing 34 distinct tasks, such as \emph{opening drawers} and \emph{pushing blocks}~\cite{MHE22}. Following~\cite{MHE22}, we adopted the ABC $\rightarrow$ D configuration, where models are trained in three different environments (A, B, and C) and evaluated in an unseen environment (D).

Each scenario consists of five consecutive tasks, evaluated using sequential success rates and the \emph{average length} metric. The former averages success rate across tasks, where failure in one task leads to failure in subsequent tasks. The latter is the sum of success rates over the five tasks, ranging from 0 to 5. We report the mean score for 100 scenarios. For baselines, we evaluated four models: GR-MG~\cite{LWH25}, MoDE~\cite{RPA24}, 3D Diffuser Actor~\cite{KGF24} and SuSIE~\cite{BNA24}.

\Cref{t:calvin} shows the results. Consistent with the findings from RLBench, integrating GraspCorrect with baseline models improved performance across evaluation tasks, with \emph{average length} enhancements ranging from 0.1 to 0.8. The gains for GR-MG and MoDE were relatively modest, which can be attributed to their already high proficiency in executing benchmark tests (see~\cref{t:calvin}). In particular, GR-MG and MoDE achieved near-perfect accuracies of 0.99 and 1.00, respectively, when only the first task was considered. Nevertheless, the results indicate that incorporating grasp pose guidance ultimately enhances execution performance in long-horizon tasks. Importantly, GraspCorrect never degraded the performance in any scenario.

\begin{table}[hbt!]
\caption{Performance on the CALVIN benchmark: Average success rates for each consecutive task and the corresponding \emph{average length} for each baseline model.}
\label{t:calvin}
\vspace{3mm}
\begin{center}
\begin{small}
\setlength{\tabcolsep}{4pt}
\begin{tabular}{l c c c c c|c} 
\toprule
 & \multicolumn{5}{c|}{\# of consecutive tasks completed} & \multicolumn{1}{c}{Avg.} \\
 \cmidrule(r){2-6}
 & 1 & 2 & 3 & 4 & 5 & len \\
\midrule
SuSIE  & 0.90 & 0.72 & 0.48 & 0.35 & 0.25 & \avglen{2.7}{1.7} \\
 + Ours & 0.96 & 0.86 & 0.72 & 0.56 & 0.41 & \avglen{3.5}{1.6} \\
\midrule
3D Diff. Act. & 0.93 & 0.79 & 0.65 & 0.55 & 0.43 & \avglen{3.4}{1.8} \\
 + Ours   & 0.97 & 0.87 & 0.77 & 0.69 & 0.54 & \avglen{3.9}{1.5} \\
\midrule
GR-MG   & 0.99 & 0.94 & 0.85 & 0.78 & 0.67 & \avglen{4.2}{1.3} \\
 + Ours   & 0.99 & 0.95 & 0.87 & 0.82 & 0.74 & \avglen{4.4}{1.3} \\
\midrule
MoDE & 1.00 & 0.94 & 0.88 & 0.81 & 0.72 & \avglen{4.4}{1.2} \\
 + Ours & 1.00 & 0.96 & 0.89 & 0.84 & 0.77 & \avglen{4.5}{1.1} \\
\bottomrule
\end{tabular}
\end{small}
\end{center}
\end{table}

\subsection{Ablation Study}
\label{s:ablation}
\paragraph{A Comparative Evaluation of PIVOT and GraspCorrect:} 
PIVOT streamlines robotic control by directly generating continuous actions from VLM outputs~\cite{NXY24}. It relies on a spatial mapping mechanism that uses camera matrices to project 3D locations onto the image plane, where grasp candidates are sampled (\cref{f:vlm}). Unlike GraspCorrect, PIVOT bypasses intermediate goal image synthesis and explicit action generation.

While PIVOT's direct approach may seem more efficient, we observed that GraspCorrect's additional processing stages significantly enhances manipulation reliability (\cref{f:vlm}). The key difference lies in GraspCorrect’s focus on grasp stability and action space exploration: PIVOT’s full 3D sampling struggles with the vast search space, often missing optimal grasp points. GraspCorrect mitigates this by leveraging egocentric views, combined with grasp-guided prompting and object-aware sampling to effectively communicate task-specific and physical constraints to the VLMs.

Our ablation studies demonstrate the impact of these core components for VLM-guided grasp detection across top five challenging tasks in RLBench (\cref{t:pivot}). Grasp-guided prompting improved the task success rate from 30.61\% to 42.52\%, while object-constrained sampling further boosted it to 73.81\%, demonstrating their combined effectiveness in achieving robust, efficient grasp detection.

Unlike full manipulation task execution, quantitatively analyzing grasp quality is challenging due to the lack of exact ground-truth data. Instead, we evaluated grasp accuracy based on the deviation of the action vector generated by GraspCorrect $\mathbf{a}^o$ (\cref{e:adecompos}) from those in successful RLBench demonstrations, treating them as (semi-)ground truth. GraspCorrect achieved an average squared Euclidean distance of 1.15, whereas it showed 1.88 before performing GraspCorrect, indicating a substantial improvement in grasp estimation (\textit{insert peg} task).

Furthermore, when considering a prediction correct if its squared distance from the ground truth is below 1.52, GraspCorrect demonstrated an average improvement of 18.75\% in accurate grasp generation.

\begin{table}[t!]
\caption{Ablation study results demonstrating the impact of grasp-guided prompting and object-aware sampling on task success rate.}
\label{t:pivot}
\renewcommand{\multirowsetup}{\centering}
\vskip 0.1in
\centering
\setlength{\tabcolsep}{5pt}
\begin{small}
{
\begin{tabular}{ccc}
\toprule
\multicolumn{2}{c}{Configuration}&\multirow{2}{*}{Success rate} \\
\cmidrule(r){1-2}
Grasp-guided prompt.&
Object-aware sampl.&
\\
\midrule
\xmark & \xmark & 30.6\\
\cmark & \xmark & 42.5 \\
\cmark & \cmark & 73.8\\
\bottomrule
\end{tabular}
}
\end{small}
\end{table}

\begin{table}[t]
\caption{Effect of action component weighting $\lambda$ (\emph{insert peg} task; RLBench).}
\label{t:weighting}
\renewcommand{\multirowsetup}{\centering}
\vskip 0.1in
\centering
\begin{small}
\setlength{\tabcolsep}{2pt}
\resizebox{\columnwidth}{!}
{
\begin{tabular}{ccccc}
\toprule
$\lambda$ & \text{1.0}&\text{0.5}&\text{0.2}&\text{0.1}\\
\midrule
\makecell{Success rate}& 78.7$\pm$16.2 & 77.3$\pm$9.2 & 82.7$\pm$2.3 & 77.3$\pm$10.1 \\
\bottomrule
\end{tabular}
}
\end{small}
\end{table}

\paragraph{Impact of Action Component Weighting $\lambda$:}
The weighting hyperparameter $\lambda$ in $\mathcal{L}$ was set to 0.2 based on RLBench validation, emphasizing the importance of orientation in effective grasping (see~\cref{t:weighting}).
As $\lambda$ deviates from this optimum value, performance degrades gracefully (\cref{t:weighting}). 

\section{Conclusion}
This work introduced GraspCorrect, a plug-and-play module designed to enhance existing robotic manipulation policies by providing precise grasping guidance. By integrating high-level semantic insights from vision-language models with detailed low-level action refinement through goal-conditioned behavioral cloning and visual goal generation, GraspCorrect significantly advances robotic manipulation capabilities. Its architecture-agnostic design enables efficient grasp refinement without requiring extensive retraining, as demonstrated across diverse manipulation tasks in RLBench and CALVIN experiments.

\paragraph{Limitations and Future Work:}
One limitation of our approach is its reliance on top-view imagery, which simplifies grasp guidance but may overlook crucial geometric features in complex 3D environments. Additionally, the framework is less effective in addressing failures that occur earlier in the manipulation pipeline or during post-grasping phases, as it primarily focuses on refining the grasp itself. Beyond these limitations, future work could explore extending the framework to handle dynamic scenes and deformable objects. Our approach relies solely on visual inputs; incorporating force and tactile feedback could further refine grasp correction allowing the system to adjust its grip based on material properties, surface friction, and object stability.

\section*{Acknowledgments}
This work was supported by the National Research Foundation of Korea (NRF) grant (No.~2021R1A2C2012195, 33\%) and the Institute of Information \& Communications Technology Planning \& Evaluation (IITP) grants (No.~RS-2019-II191906, AI Graduate School Program, POSTECH, 33\%; and No.~RS-2022-II220290, Visual Intelligence for Space-Time Understanding and Generation, 33\%), funded by the Korean government (MSIT).

\bibliography{GraspCorrect}
\bibliographystyle{icml2025}

\newpage
\appendix
\onecolumn

\section{Implementation and Experimental Details}
\label{s:implementationdetails}
\subsection{VLM-guided Grasp Detection}
\label{s:vlmguidedgraspdetails}
\paragraph{Grasp-Guided Prompts:}
\label{sss:prompt}
\let\tqs\textquotesingle
\begin{center}
\fbox{\parbox{0.97\linewidth}
{\footnotesize{
\texttt{You are a robot equipped with a parallel-jaw gripper, performing the task \tqs\{task\_desc\}\tqs. Analyze the provided pre-grasp pose of an object and specify precise contact positions for each robot gripper to achieve a stable grasp. Describe the contact position as much detail as possible using numerical expressions. Avoid using exact coordinates. Respond in the format: \tqs Left: [1 sentence starting with \tqs\tqs Position the left gripper\tqs\tqs]. Right: [1 sentence starting with \tqs\tqs Position the right gripper\tqs\tqs]\tqs. Let's think step by step.}
}}}
\end{center}

Every RLBench task has various textual descriptions \texttt{task\_desc} (e.g., ``put the ring on the maroon spoke'', ``open the bottom drawer'', ``place 1 cup on the cup holder'') to be used for linguistic learning in robotic tasks.

\paragraph{Iterative VQA:}
\begin{center}
\fbox{\parbox{0.97\linewidth}
{\footnotesize{
\texttt{INSTRUCTIONS: You are tasked to locate an object, region, or point in space in the given annotated image according to a description. The image is annotated with numbered circles. \\
Choose the top \{top\_n\} circles that have the most overlap with and/or is closest to what the description is describing in the image.
You are a five-time world champion in this game. Give a one sentence analysis of why you chose those points. Provide your answer at the end in a valid JSON of this format:
\text{{{"points": []}}}. \\
DESCRIPTION: \{description\} \\
IMAGE: \{image\}
}}}}
\end{center}

Grasp candidates are chosen by iteratively querying VLM using the above prompt adapted from PIVOT~\cite{NXY24}'s GitHub repository. In this process, we set \texttt{top\_n} value to 3, while the \texttt{description} corresponds to the output generated during the grasp-guided prompting phase.

Our module introduces a sequential approach to parallel-jaw gripper positioning that significantly improves spatial coherence during the VQA process. This method begins by determining the left gripper's positioning, marking it on the input image, and then incorporating this information through additional language prompts (``\texttt{Be aware that the red circle indicates the left gripper's contact position}'') when determining the right gripper's position. We find that independent positioning which determines contact points for both grippers in isolation frequently results in suboptimal or physically infeasible grasp configurations, primarily due to the lack of inter-gripper spatial awareness. By explicitly incorporating the first gripper's position into the decision process for the second gripper, our sequential method enables the VLM to maintain comprehensive spatial awareness throughout the grasp correction process.

\subsection{Action Generation}
\label{ss:action}
\paragraph{Model Architecture:}
The policy network architecture consists of a ResNet-34~\cite{HZR16} with Group Normalization~\cite{WH18}, which processes the current and goal observation images stacked along the channel dimension. The encoder's output is then passed through a 3-layer Multi-Layer Perceptron (MLP) with 256 hidden units and Swish activations~\cite{HG16} in each layer. This MLP outputs the mean and standard deviation for a Gaussian action distribution.

Training is performed using the Adam optimizer~\cite{KB14} with a learning rate of 5e-4, a linear warmup schedule over 2,000 steps, and a batch size of 256. Prior to concatenation, both current and goal images undergo standard image augmentations, including random cropping, as well as brightness, contrast, saturation, and hue adjustments.

\paragraph{Data Generation Protocol:}
\begin{figure}[hbt!]
\begin{center}
\includegraphics[width=0.8\columnwidth]{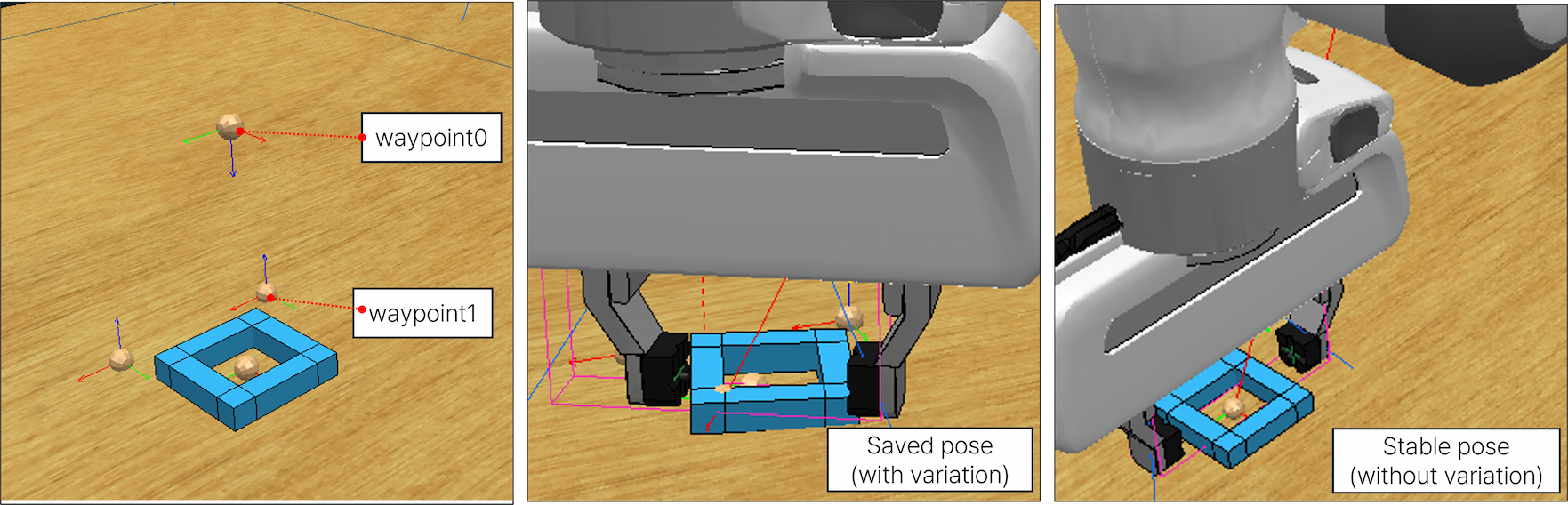}
\caption{Data generation process for policy training. Left: waypoints are randomly varied to introduce realistic grasping variations. Middle: the resulting grasp pose with variation is saved. Right: the grasp pose without waypoint variation represents the stable grasp.}
\label{f:data}
\end{center}
\end{figure}
Our action generation policy requires paired data consisting of observation-action tuples ($\mathbf{o}_{t(g)}$, $\mathbf{a}_{t(g)}$, $\mathbf{o}_{t(g)}^{*}$, $\mathbf{a}_{t(g)}^{*}$), where states requiring grasp correction are paired with their corresponding stable grasp configurations. We implement a two-stage data collection protocol within the RLBench environment to generate these training pairs.

For collecting correction-needed states ($\mathbf{o}_{t(g)}$, $\mathbf{a}_{t(g)}$), we first initialize a simulation environment comprising a tabletop workspace, target object, and Franka Panda robot. The motion path is defined by waypoints (\cref{f:data} left) that serve as reference points for RLBench's path planning algorithm. By introducing controlled randomization to these waypoint positions and orientations, we generate realistic variations in grasp attempts. At the moment of grasping, we record both the observation and the executed action vector. To obtain stable reference states ($\mathbf{o}_{t(g)}^{*}$, $\mathbf{a}_{t(g)}^{*}$), we repeat the grasping sequence under identical conditions but without waypoint randomization. Through this systematic process, we generate 200 paired examples for each manipulation task, providing a balanced dataset for policy learning.
\newpage
\section{Visualization}
\begin{figure}[hbt!]
\vspace{3mm}
\begin{center}
\includegraphics[width=0.75\columnwidth]{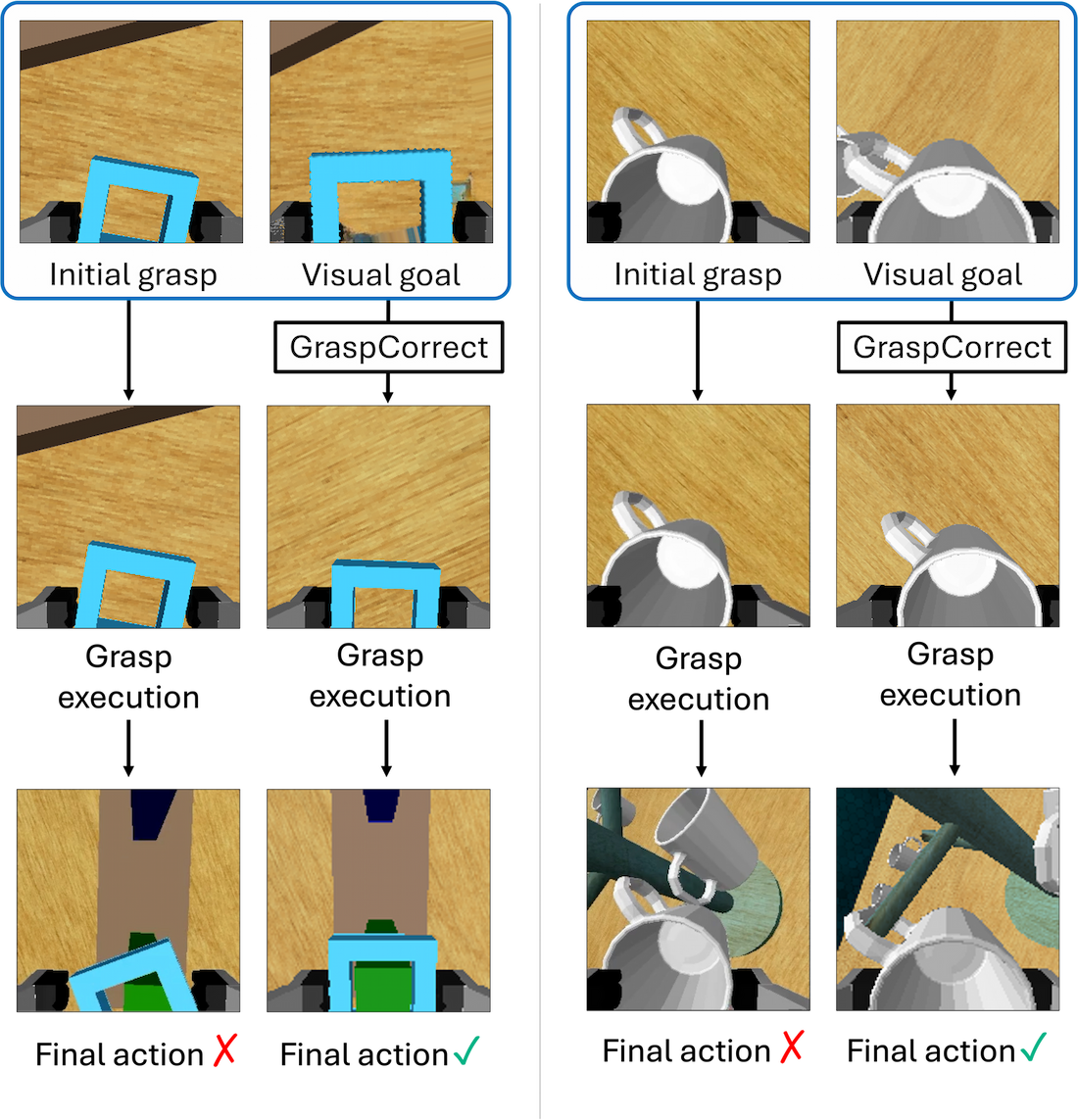}
\caption{Visualization of manipulation processes with and without GraspCorrect. (Left) \emph{insert peg} and (Right) \emph{place cups}. The GraspCorrect-assisted manipulation process consists of four stages: initial grasp pose, visual goal generation, actual grasp execution, and final action. By effectively correcting initial grasp inaccuracies, GraspCorrect ensures successful task completion.}
\label{f:visualization}
\end{center}
\end{figure}

\begin{figure}[hbt!]
    \vspace{3mm}
    \begin{center}
        \setlength{\tabcolsep}{1.5pt}
        \begin{tabular}{cccc}
            \includegraphics[width=0.235\columnwidth]{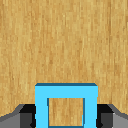} &
            \includegraphics[width=0.235\columnwidth]{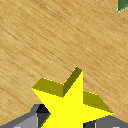} &
            \includegraphics[width=0.235\columnwidth]{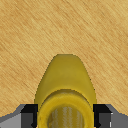} &
            \includegraphics[width=0.235\columnwidth]{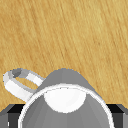} \\
            \includegraphics[width=0.235\columnwidth]{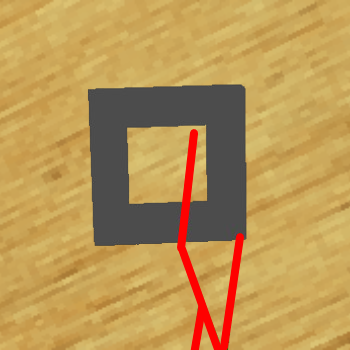} & 
            \includegraphics[width=0.235\columnwidth]{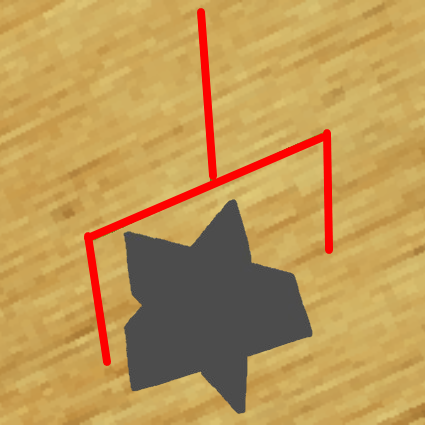} & 
            \includegraphics[width=0.235\columnwidth]{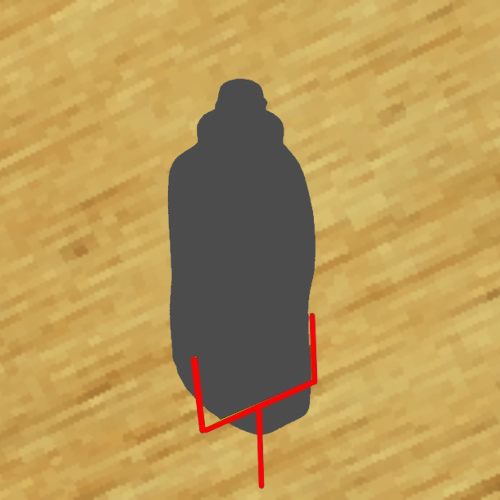} & 
            \includegraphics[width=0.235\columnwidth]{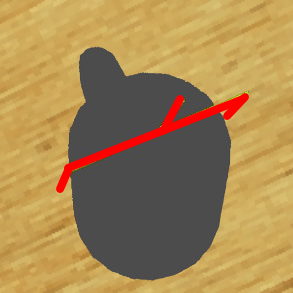} \\
            \includegraphics[width=0.235\columnwidth]{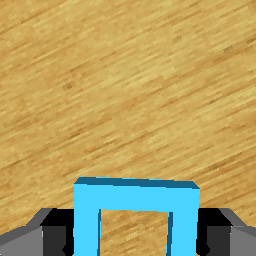} & 
            \includegraphics[width=0.2355\columnwidth]{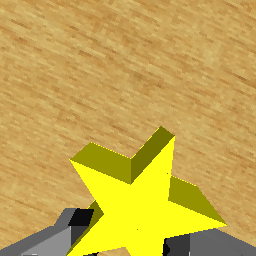} & 
            \includegraphics[width=0.235\columnwidth]{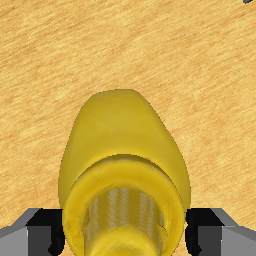} & 
            \includegraphics[width=0.235\columnwidth]{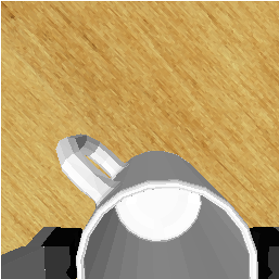} \\
        \end{tabular}
        \caption{Comparison of grasp poses across four manipulation tasks: \textit{insert peg}, \textit{sort shape (star)}, \textit{put in cupboard (mustard)}, and \textit{place cups}; Top: ground-truth grasps from successful manipulations in RLBench. Middle: Contact-GraspNet (red markers indicate the predicted gripper poses). Bottom: ours.}
        \label{f:graspexample}
    \end{center}
\end{figure}
\end{document}